\documentclass{article} 
\usepackage{iclr2018_conference,times}
\usepackage{hyperref}
\usepackage{url}

\usepackage{amsmath}
\usepackage{times}
\usepackage{graphicx} 
\usepackage{subfigure} 
\usepackage{color}

\usepackage{algorithm}
\usepackage[noend]{algpseudocode}
\usepackage{hyperref}
\usepackage{verbatim}
\usepackage{sidecap, caption}
\usepackage{multicol}

\newcommand{\fig}[1]{Fig.~\ref{fig:#1}}
\newcommand{\tab}[1]{Table~\ref{tab:#1}}

\newcommand{\append}[1]{Appendix~\ref{app:#1}}

\newcommand{\tmax}{t_\text{Max}}

\title{Intrinsic Motivation and Automatic \\ Curricula via Asymmetric Self-Play}

\author{Sainbayar Sukhbaatar \\
Dept. of Computer Science\\
New York University\\
\texttt{sainbar@cs.nyu.edu} \\
\And
Zeming Lin\\
Facebook AI Research\\
New York\\
\texttt{zlin@fb.com} \\
\And
Ilya Kostrikov \\
Dept. of Computer Science\\
New York University\\
\texttt{kostrikov@cs.nyu.edu} \\
\And
Gabriel Synnaeve, Arthur Szlam \& Rob Fergus\\
Facebook AI Research\\
New York\\
\texttt{\{gab,aszlam,robfergus\}@fb.com} \\
}

\iclrfinalcopy 

\begin{document}

\maketitle

\begin{abstract} 
  We describe a simple scheme that allows an agent to learn about its
  environment in an unsupervised manner. Our scheme pits two versions
  of the same agent, Alice and Bob, against one another. Alice
  proposes a task for Bob to complete; and then Bob attempts to
  complete the task.  In this work we will focus on two kinds of environments: (nearly)
  reversible environments and environments that can be reset.
  Alice will ``propose'' the task by doing a sequence of actions and then
  Bob must undo or repeat them, respectively.  Via an
  appropriate reward structure, Alice and Bob automatically generate a
  curriculum of exploration, enabling unsupervised training of the
  agent. When Bob is deployed on an RL task within the environment, this
  unsupervised training reduces the number of supervised episodes needed to
  learn, and in some cases converges to a higher reward.
\end{abstract} 

\section{Introduction}
Model-free approaches to reinforcement learning are sample inefficient,
typically requiring a huge number of episodes to learn a satisfactory
policy. The lack of an explicit environment model means the agent must
learn the rules of the environment from scratch at the same time as it tries to understand which
trajectories lead to rewards. In environments where reward is sparse,
only a small fraction of the agents' experience is directly used to
update the policy, contributing to the inefficiency.

In this paper we introduce a novel form of unsupervised training
for an agent that enables exploration and learning about the
environment without any external reward that incentivizes the agents to learn
how to transition between states as efficiently as possible.
We demonstrate that this unsupervised training allows the agent
to learn new tasks within the environment quickly.


\begin{figure}[h!]
\begin{center}
\includegraphics[width=0.9\columnwidth]{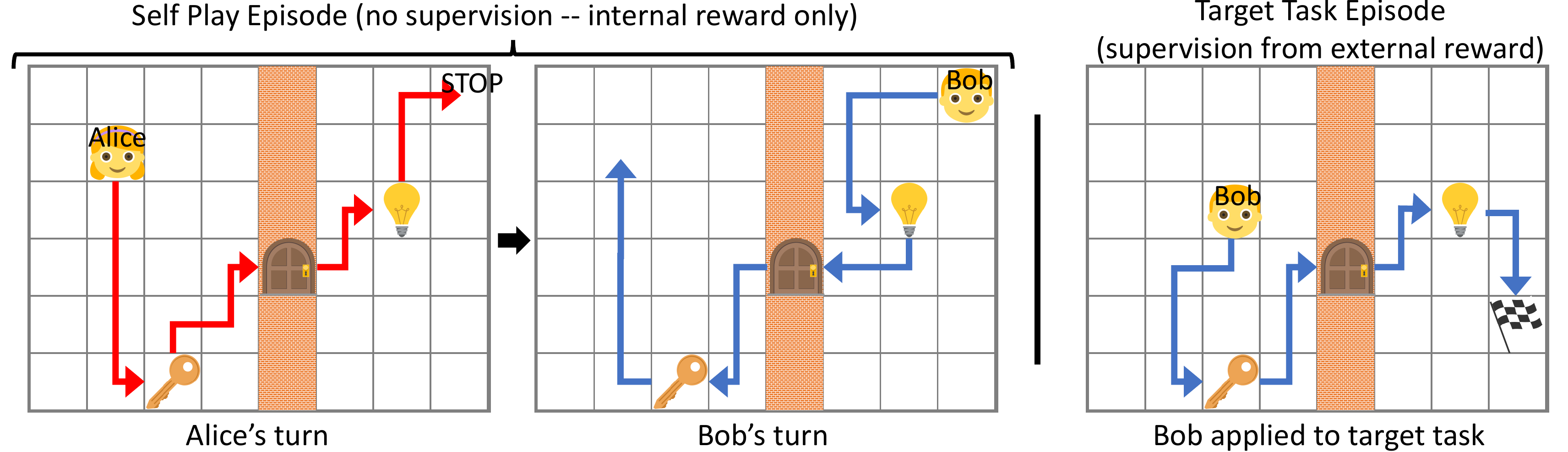}
\vspace{-3mm}
\caption{ Illustration of the self-play concept in a gridworld
  setting. Training consists of two types of episode: self-play and
  target task. In the former, Alice and Bob take turns moving the
  agent within the environment. Alice sets tasks by altering the state
  via interaction with its objects (key, door, light) and then hands
  control over to Bob. He must return the environment to its original
  state to receive an internal reward. This task is just one of many
  devised by Alice, who automatically builds a curriculum of
  increasingly challenging tasks. In the target task, Bob's policy is
  used to control the agent, with him receiving an {\em external}
  reward if he visits the flag. He is able to learn to do this quickly
  as he is already familiar with the environment from self-play.}
\label{fig:mazebase}
\end{center}
\end{figure}

\section{Approach}
\vspace{-1.5mm}
We consider environments with a single physical agent 
(or multiple physical units controlled by a single agent), but we allow it
to have two separate ``minds'': Alice and Bob, each with its own
objective and parameters. During self-play episodes, Alice's job is to propose
a task for Bob to complete, and Bob's job is to complete the
task. When presented with a target task episode, Bob is then used to perform
it (Alice plays no role). The key idea is that the Bob's play with
Alice should help him understand how the environment works and
enabling him to learn the target task more quickly.

Our approach is restricted to two classes of environment: (i)
those that are (nearly) reversible, or (ii) ones that can be reset to
their initial state (at least once). These restrictions allow us to
sidestep complications around how to communicate the task and
determine its difficulty (see \append{comm}
for further discussion).
In these two scenarios, Alice starts at some initial state $s_0$ and
proposes a task by {\it doing} it, i.e. executing a sequence of
actions that takes the agent to a state $s_t$. She then outputs a {\sc
  stop}
action, which hands control over to Bob. In reversible environments,
Bob's goal is to return the agent back to state $s_0$ (or within some
margin of it, if the state is continuous), to receive reward. In partially observable
environments, the objective is relaxed to Bob finding a state that returns the same observation as Alice's
initial state. In environments where resets are permissible, Alice's
{\sc stop} action
also reinitializes the environment, thus Bob starts at $s_0$ and now
must reach $s_t$ to be rewarded, thus repeating Alice's task instead
of reversing it. See \fig{mazebase} for an example,
and also Algorithm 1 in \append{code}. 

In both cases, this self-play between Alice and Bob only involves
internal reward (detailed below), thus the agent can be
trained without needing any supervisory signal from the
environment. As such, it comprises a form of unsupervised training where
Alice and Bob explore the environment and learn how it operates. This
exploration can be leveraged for some target task by 
training Bob on target task episodes in parallel.
The idea is that Bob's experience from self-play 
will help him learn the target task in fewer episodes.
The reason behind choosing Bob for the target task is because
he learns to transfer from one state to another efficiently from self-play. 
See Algorithm~2 in \append{code} 
 for detail.

For self-play, we choose the reward structure for Alice and Bob to encourage
Alice to push Bob past his comfort zone, but not give him impossible
tasks. Denoting Bob's total reward by $R_B$ (given at the end of episodes) 
and Alice's total reward
by $R_A$, we use
\begin{equation}\label{eq:rb}
 R_B = -\gamma t_B
\end{equation}
where $t_B$ is the time taken by Bob to complete his task and 
\begin{equation}\label{eq:ra}
R_A = \gamma \max(0,t_B - t_A)
\end{equation}
 where $t_A$ is the time until Alice
performs the {\sc stop}
action, and $\gamma$ is a scaling coefficient that balances this
internal reward to be of the same scale as external rewards from the target task. 
The total length of an episode is limited to $\tmax$, so if Bob fails to complete the task in time we set $t_B = \tmax - t_A$.

Thus Alice is rewarded if Bob takes more time, but the
negative term on her own time will encourage Alice not to take too
many steps when Bob is failing. 
For both reversible and resettable environments, Alice must limit
her steps to make Bob's task easier, thus Alice's optimal behavior is to the find simplest
tasks that Bob cannot complete. This eases learning for Bob since the
new task will be only just beyond his current
capabilities. The self-regulating feedback between Alice and Bob allows them to
automatically construct a curriculum for exploration, a key
contribution of our approach. 

\vspace{-2mm}
\subsection{Parameterizing Alice and Bob's actions}
Alice and Bob each have  policy functions which take
as input two observations of state variables, and output a distribution over
actions. In Alice's case, the function will be of the form
\[a_{\text{A}} = \pi_A(s_t,s_0),\] where $s_0$ is the observation of
the initial state of the environment and $s_t$ is the observation of
the current state. In Bob's case, the function will be
\[a_{\text{B}} = \pi_B(s_t,s^*),\] where $s^*$ is the target state
that Bob has to reach, and set to $s_0$ when we have a
reversible environment. In a resettable environment $s^*$ is the state where Alice executed
the STOP action. 

When a target task is presented, the agent's policy
function is $a_{\text{Target}} = \pi_B(s_t,\emptyset),$
where the second argument of Bob's policy is simply set to zero 
\footnote{Note that Bob can be used in multi-task learning by feeding the task description into $\pi_B$}. 
If $s^*$ is always non-zero, then this is enough to let Bob know
whether the current episode is self-play or target task.
In some experiments where $s^*$ can be zero, we give third argument 
$z \in \{0,1\}$ that explicitly indicates the episode kind.

In the experiments below, we demonstrate our approach in settings where
$\pi_A$ and $\pi_B$ are tabular; where it is a neural network taking discrete inputs,
and where it is a neural network taking in continuous inputs.
When using a neural network, we use the same network architecture 
for both Alice and Bob, except they have different parameters
\[\pi_A(s_t,s_0)=f(s_t,s_0,\theta_A), \quad \pi_B(s_t,s^*) = f(s_t,s^*,\theta_B) ,\]
where $f$ is an multi-layered neural network with parameters $\theta_A$ or $\theta_B$.

\subsection{Universal Bob in the tabular setting}
\label{sec:universal}
We now present a theoretical argument that shows for environments with
finite states, tabular policies, and deterministic, Markovian transitions, we can
interpret the self-play as training Bob to find
a policy that can get from any state to any other in the least
expected number of steps.

{\bf Preliminaries:} Note that, as discussed above, the policy table
for Bob is indexed by $(s_t,s^*)$, not just by $s_t$.  In
particular, with the assumptions above, this means that there is a
{\em fast policy} $\pi_{\text{fast}}$ such
that $\pi_{\text{fast}}(s_{t},s^*)$ has
the smallest expected number of steps to transition from
$s_{t}$ to $s^*$.    It is clear that $\pi_{\text{fast}}$ is a universal policy for Bob,
such that $\pi_B=\pi_{\text{fast}}$ is optimal
with respect to any Alice's policy $\pi_A$.
In a reset game, $\pi_{\text{fast}}$
nets Alice a return of 0, and in the reverse game, the return of
$\pi_{\text{fast}}$ against an optimal Alice  can be considered a measure of the reversibility
of the environment. However, in what follows let us assume that either
the reset game or the reverse game in a perfectly reversible
environment is used. Also, let assume the initial states are 
randomized and its distribution covers the entire state space.

{\bf Claim:} If $\pi_A$ and $\pi_B$ are policies of Alice and Bob that
are in equilibrium (i.e., Alice cannot be made better without changing
Bob, and vice-versa), then $\pi_B$ is a fast policy. 

{\bf Argument:} 
Let us first show that Alice will always get zero reward in equilibrium.
If Alice is getting positive reward on some challenge, 
that means Bob is taking longer than Alice on that challenge.
Then Bob can be improved to use $\pi_{fast}$ at that challenge, which contradicts the equilibrium assumption.

Now let us prove $\pi_B$ is a fast policy by contradiction. 
If $\pi_B$ is not fast, then there must exist a challenge $(s_t, s^*)$
where $\pi_B$ will take longer than $\pi_{fast}$. 
Therefore Bob can get more reward by using $\pi_{fast}$ 
if Alice does propose that challenge with non-zero probability.
Since we assumed equilibrium and $\pi_B$ cannot be improved while $\pi_A$ fixed, 
the only possibility is that Alice is never proposing that challenge.
If that is true, Alice can get positive reward by proposing that task
using the same actions as $\pi_{fast}$, so taking fewer steps than $\pi_B$. 
However this contradicts with the proof that Alice always gets zero reward, making our 
initial assumption ``$\pi_B$ is not fast'' wrong.

\section{Related Work}
\vspace{-2mm}

Self-play arises naturally in reinforcement learning, and has been well studied.  
For
example, for playing checkers \citep{DBLP:journals/ibmrd/Samuel59}, backgammon \citep{DBLP:journals/cacm/Tesauro95}, and Go, \citep{Silver_2016_go}, and 
in multi-agent games 
such as
RoboSoccer \citep{riedmiller2009reinforcement}. Here, the agents or teams of
agents compete for external reward.  This differs from
our scheme where the reward is purely internal and the self-play
is a way of motivating an agent to learn about its environment to augment sparse rewards from separate target
tasks.

Our approach has some relationships with 
generative adversarial
networks (GANs) \citep{goodfellow2014generative}, which train a generative neural net by having
it try to fool a discriminator network which tries to differentiate
samples from the training examples. 
\cite{Li17} introduce an
adversarial approach to dialogue generation, where a generator model
is subjected to a form of ``Turing test'' by a discriminator
network. \cite{Mescheder17} demonstrate how adversarial loss terms can
be combined with variational auto-encoders to permit more accurate
density modeling.   While GAN's are often thought of as methods for training a generator, the generator can be thought of as a method for generating hard negatives for the discriminator.  From this viewpoint, 
in our approach, Alice acts as a ``generator'', finding ``negatives'' for Bob. 
However, Bob's jobs is to complete the generated challenge, not to discriminate it.

There is a large body of work on intrinsic motivation \citep{Barto2013,intrinsically_motivated_RL,empowerment, Schmidhuber-ijcnn-91} for self-supervised learning agents.  These works propose methods for training an agent to explore and become proficient at manipulating its environment without  necessarily having a specific target task, and without a source of extrinsic supervision.   
One line in this direction is curiosity-driven exploration \citep{Schmidhuber-ijcnn-91}.  These techniques can be applied in encouraging exploration in the context of reinforcement learning, for example \citep{unifying_exploration, mbie, model_based_exploration, Tang16,pathakICMl17curiosity}; 
Roughly, these use some notion of the novelty of a state to give a reward.  In the simplest
setting, novelty can be just the number of times a state has been visited; in more complex scenarios,
the agent can build a model of the world, and the novelty is the difficulty in placing the current state into the model.  In our work, there is no explicit notion of novelty.  Even if Bob has seen a state many times, if he has trouble getting to it, Alice should force him towards that state.
Another line of work on intrinsic motivation is a formalization of the notion of empowerment \citep{empowerment}, or how much control the agent has over its environment.   Our work is related in the sense that it is in both Alice's and Bob's interests to have more control over the environment; but we do not explicitly measure that control except in relation to the tasks that Alice sets.


Curriculum learning \citep{bengio2009curriculum} is widely used in many
machine learning approaches. Typically however, the curriculum
requires at least some manual specification. A key point about our
work is that Alice and Bob devise their own curriculum entirely
automatically. Previous automatic approaches, such as \cite{Kumar10},
rely on monitoring training error. But since ours is unsupervised, no
training labels are required either.

Our basic paradigm of ``Alice proposing a task, and Bob doing it'' is related to the Horde architecture 
\citep{horde} and \citep{uvfa}.  In those works, instead of using a value function $V = V(s)$ that depends on 
the current state, a value function that explicitly depends on state and goal $V = V(s,g)$ is used.  In our experiments, our models will be parameterized in a similar fashion.  The novelty in this work is in how Alice defines the goal for Bob.

The closest work to ours is that of 
\cite{Baranes13}, who also have one part of the model that proposes
tasks, while another part learns to complete them.  As in this work, the policies and cost are
parameterized as functions of both state and
goal. However, our approach differs in the way tasks are proposed and communicated.   In particular, 
in \cite{Baranes13}, the goal space has to be presented in a way that allows explicit partitioning and sampling, 
whereas in our work, the goals are sampled through Alice's actions.  On the other hand, we pay for not having
to have such a representation by requiring the environment to be either reversible or resettable.

Several concurrent works are related: \cite{Andrychowicz17}
form an implicit curriculum by using internal states as
a target. \cite{Florensa17a} automatically generate a
series of increasingly distant start states from a
goal. \cite{Pinto17} use an adversarial framework to perturb the
environment, inducing improved robustness of the agent.  \cite{Held17} propose a scheme related to our ``random
Alice'' strategy\footnote{In their paper they analyzed our approach,
suggesting it was inherently unstable. However, the analysis relied on
a sudden jump of Bob policy with respect to Alice's, which is unlikely
to happen in practice.}. 



\vspace{-2mm}
\section{Experiments}
\vspace{-2mm} The following experiments explore our self-play approach
on a variety of tasks, both continuous and discrete, from the Mazebase
\citep{mazebase}, RLLab \citep{duan2016benchmarking}, and
StarCraft\footnote{StarCraft is a trademark or registered trademark of Blizzard Entertainment, Inc., in the U.S. and/or other countries.  Nothing in this paper should not be construed as approval, endorsement, or sponsorship by Blizzard Entertainment, Inc.} environments~\citep{synnaeve2016torchcraft}.  The same
protocol is used in all settings: self-play and target task episodes
are mixed together and used to train the agent via discrete policy
gradient. We evaluate both the reverse and repeat versions of
self-play.  We demonstrate that the self-play episodes help training,
in terms of number of target task episodes needed to learn the
task. Note that we assume the self-play episodes to be ``free'', since
they make no use of environmental reward. This is consistent with
traditional semi-supervised learning, where evaluations typically are based only
on the number of labeled points (not unlabeled ones too).

In all the experiments we use policy gradient \citep{Williams92simplestatistical} with a baseline
for optimizing the policies.  In the tabular task below, we use a constant baseline; in all the other tasks
we use a policy parameterized by a neural network, and a baseline that depends on the state. We denote the
states in an episode by $s_1,...,s_T$, and the actions taken at each
of those states as $a_1,...,a_T$, where $T$ is the length of the
episode. The baseline is a scalar function of the states
$b(s,\theta)$, computed via an extra head on the network producing the
action probabilities. Besides maximizing the expected reward with
policy gradient, the models are also trained to minimize the distance
between the baseline value and actual reward. Thus after finishing
an episode, we update the model parameters $\theta$ by
\[
\Delta \theta = \sum_{t=1}^T \left[ 
  \frac{\partial \log f(a_t| s_t, \theta)}{\partial \theta} 
  \left(\sum_{i=t}^T r_i - b(s_t, \theta) \right) \right.
   \left.- \lambda
  \frac{\partial}{\partial \theta} 
  \left(\sum_{i=t}^T r_i - b(s_t, \theta) \right)^2 
\right] .
\label{eq:RL}
\]
Here $r_t$ is reward given at time $t$, and the hyperparameter $\lambda$ is for balancing the reward and the baseline objectives, which is set to 0.1 in all experiments. 

For the policy neural networks, we use two-layer fully-connected networks with 50 hidden units in each layer.
The training uses RMSProp \citep{rmsprop}. We always do 10 runs with different random initializations
and report their mean and standard deviation.
See \append{hyper}
for all the hyperparameter values used in the experiments.

\vspace{-2mm}
\subsection{Long hallway}
\label{sec:hallway}
We first describe a simple toy environment designed to illustrate the function of
the asymmetric self-play.   The environment consists of $M$ states
$\{s_1,...,s_M\}$ arranged in a chain.  Both Alice and Bob have three
possible actions, ``left'', ``right'', or ``stop''.  If the agent is
at $s_i$ with $i\neq 1$, ``left'' takes it to $s_{i-1}$; ``right''
analogously increases the state index, and ``stop'' transfers control
to Bob when Alice runs it and terminates the episode when Bob runs it.
We use ``return to initial state'' as the self-play task (i.e. Reverse
in Algorithm 1 in \append{code}
).  For the
target task, we randomly pick a starting state and target state, and the
episode is considered successful if Bob moves to the target state and
executes the stop action before a fixed number of maximum steps. 

In this case, the target task is essentially the same as the self-play
task, and so running it is not unsupervised learning (and in particular, on this toy example
unlike the other examples below, we do not mix self-play training with
target task training). 
 However, we see
that the curriculum afforded by the self-play is efficient at training
the agent to do the target task at the beginning of the training, and is
effective at forcing exploration of the state space as Bob gets more
competent.

In \fig{hallway_mazebase} (left) we plot the number of episodes vs rate of success at
the target task with four different methods.  We set $M=25$ and the
maximum allowed steps for Alice and Bob to be $30$.  We use fully
tabular controllers; the table is of size $M^2 \times 3$, with a
distribution over the three actions for each possible (start, end
pair).

The red curve corresponds to policy gradient, with a penalty of
$-1$ 
given upon failure to complete the
task, and a penalty of $-t/\tmax$ for successfully completing the task in $t$ steps.  The magenta curve corresponds to taking Alice
to have a random policy ($1/2$ probability of moving left or right,
and not stopping till the maximum allowed steps).  The green curve
corresponds to policy gradient with an exploration bonus similar to
\cite{mbie}.  That is, we keep count of the number of times $N_s$ the
agent has been in each state $s$, and the reward for $s$ is adjusted
by exploration bonus $\alpha/\sqrt{N_s}$, where $\alpha$ is a constant
balancing the reward from completing the task with the exploration
bonus.  We choose the weight $\alpha$ to maximize success at 0.2M
episodes from the set $\{0, 0.1, 0.2,...,1\}$.  The blue curve
corresponds to the asymmetric self-play training.
 
We can see that at the very beginning, a random policy for Alice gives
some form of curriculum but eventually is harmful, because Bob never
gets to see any long treks.  On the other hand, policy gradient sees
very few successes in the beginning, and so trains slowly.  Using the
self-play method, Alice gives Bob easy problems at first (she
starts from random), and then builds harder and harder problems as the
training progresses, finally matching the performance boost of the
count based exploration. Although not shown, similar patterns are
observed for a wide range of learning rates.

\vspace{-2mm}
\subsection{Mazebase: Light key}
We now describe experiments using the MazeBase environment
\citep{mazebase}, which have discrete actions and states,
but sufficient combinatorial complexity that tabular
methods cannot be used.  The environment consist of various items placed on a finite
2D grid; and randomly generated for each episode.

We use an environment where
the maze contains a light switch (whose initial state is sampled
according to a predefined probability, p(Light off)), a
key and a wall with a door (see \fig{mazebase}). An agent can open or
close the door by toggling the key switch, and turn on or off light
with the light switch. When the light is off, the agent can only see the
(glowing) light switch. In the target task, there is also a goal flag item,
and the objective of the game is reach to that goal flag.

In self-play, the environment is the same except there is no specific objective.
An episode starts with Alice in control, who can
navigate through the maze and change the switch states until she
outputs the STOP action. Then, Bob takes control and tries to return
everything to its original state (restricted to visible items)
in the reverse self-play. 
In the repeat version, the maze resets back to its initial state 
when Bob takes the control, who tries to reach the final state of Alice.

\begin{figure}[h!]
\begin{center}
\mbox{
\includegraphics[width=0.40\columnwidth]{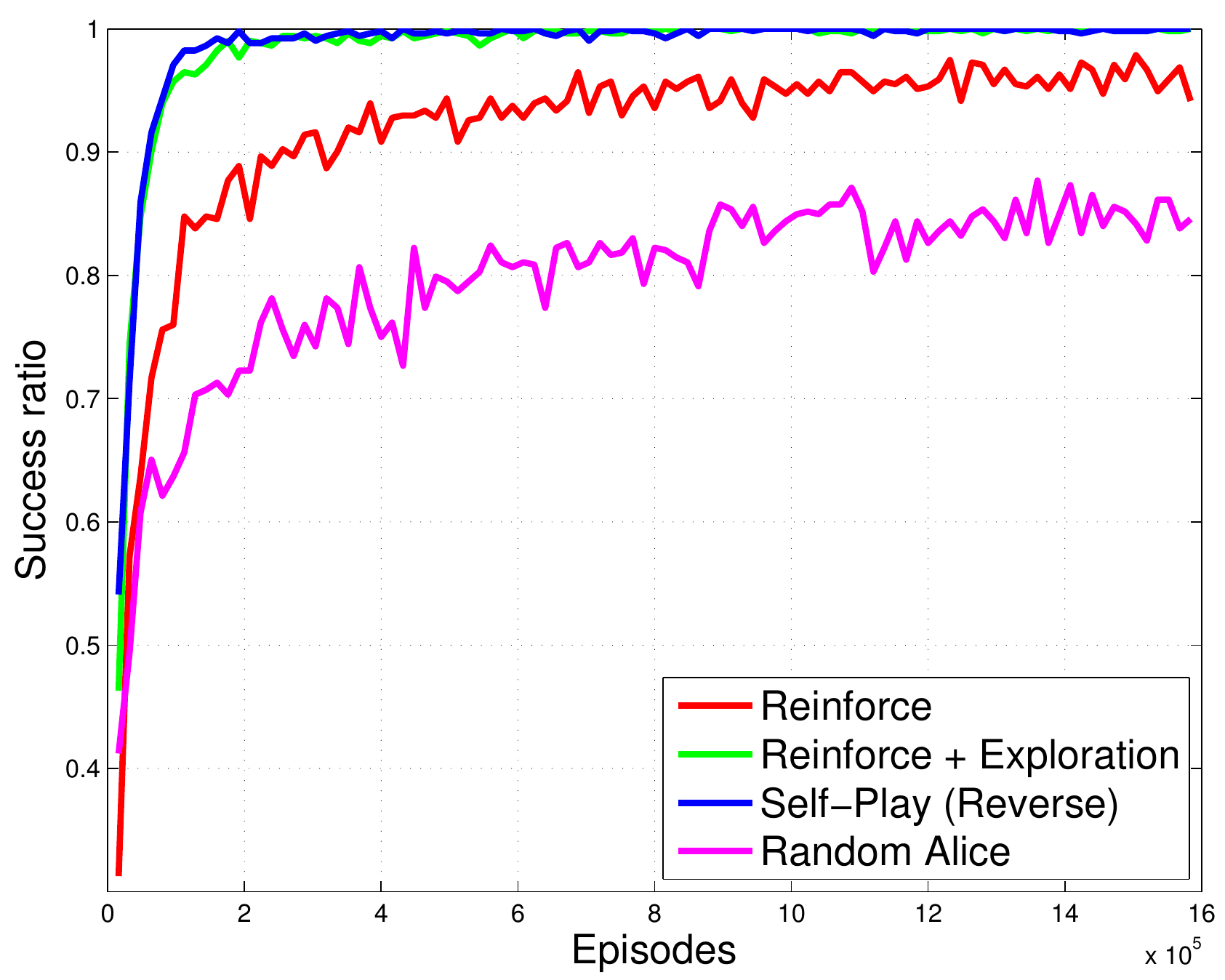}
\hspace{6mm}
\includegraphics[width=0.40\columnwidth]{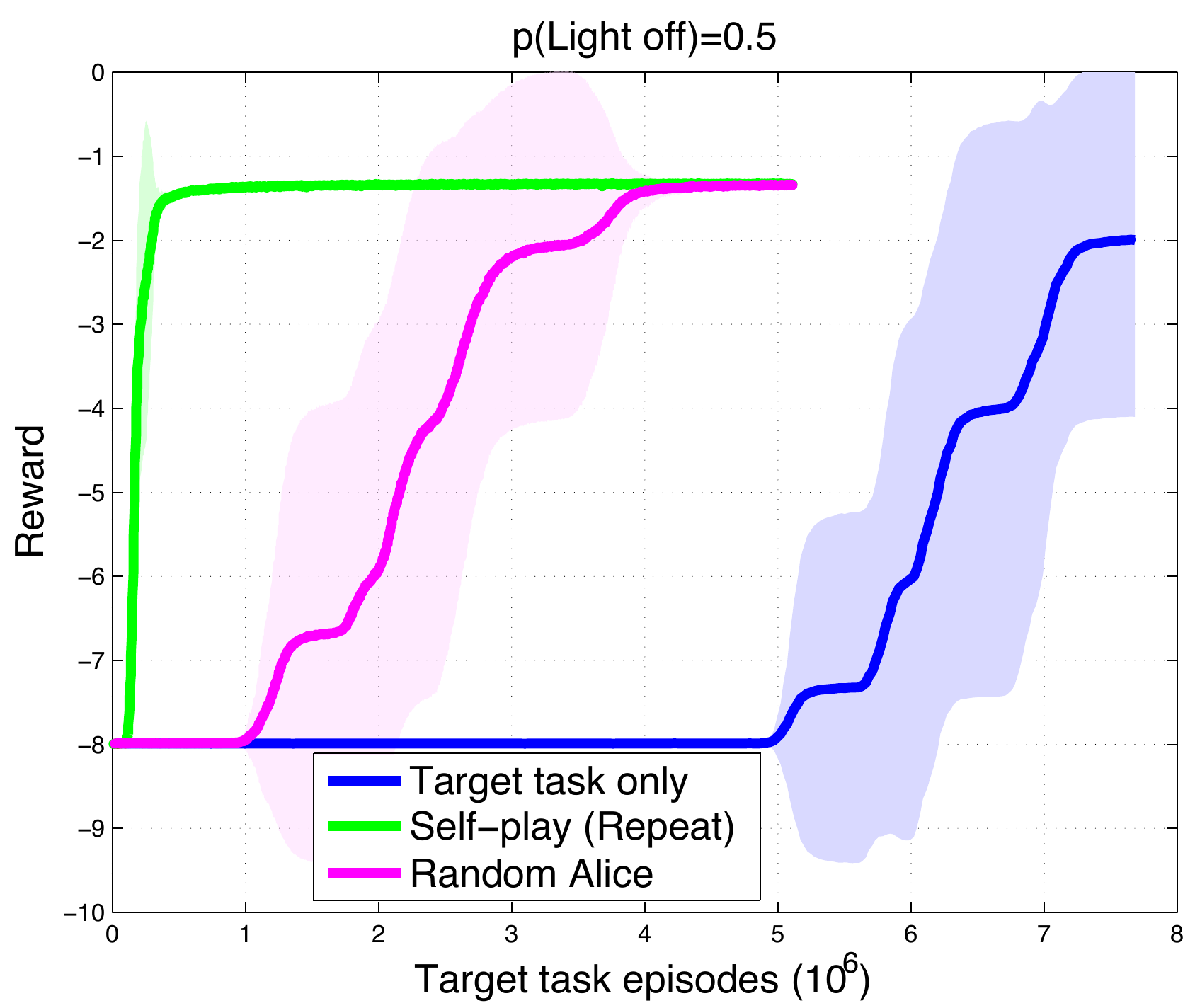}
}
\caption{\textbf{Left:} The hallway task from section \ref{sec:hallway}.  The $y$
  axis is fraction of successes on the target task, and the $x$ axis is
  the total number of training examples seen.  Standard policy gradient (red) learns slowly. Adding
  an explicit exploration bonus \citep{mbie} (green) helps significantly. Our self-play
  approach (blue) gives similar performance however. Using a random
  policy for Alice (magenta) drastically impairs performance, showing
  the importance of self-play between Alice and Bob. 
  \textbf{Right:} Mazebase task, illustrated in \fig{mazebase}, for p(Light
  off) = 0.5. Augmenting with the repeat form of self-play enables
  significantly faster learning than training on the target task alone
  and random Alice baselines. } 
\label{fig:hallway_mazebase}
\end{center}
\end{figure}
\begin{figure}[t!]
\begin{center}
\includegraphics[width=1\columnwidth]{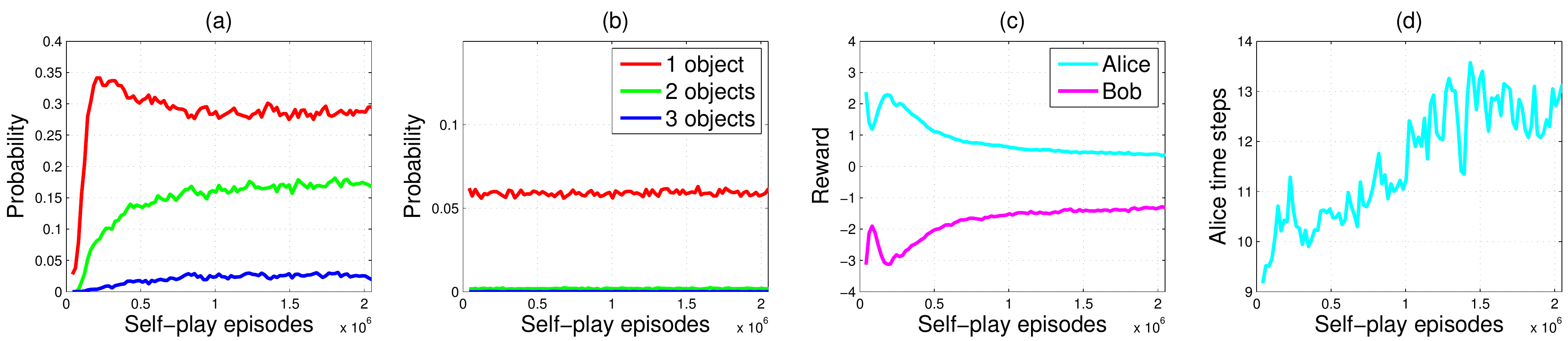}
\vspace{-4mm}
\caption{Inspection of a Mazebase learning run, using the 
  environment shown in \fig{mazebase}. (a): rate at which
  Alice interacts with 1, 2 or 3 objects during an
  episode, illustrating the automatically generated curriculum. Initially
  Alice touches no objects, but then starts to interact with one. But
  this rate drops as Alice devises tasks that involve two and subsequently three objects. (b) by contrast, in the
  random Alice baseline, she never utilizes more than a single object
  and even then at a much lower rate. (c) plot of Alice and Bob's
  reward, which strongly correlates with (a). (d) plot of $t_a$ as self-play progresses. Alice takes an
  increasing amount of time before handing over to Bob, consistent
  with tasks of increasing difficulty being set.} 
\vspace{-5mm}
\label{fig:4}
\end{center}
\end{figure}

In \fig{hallway_mazebase} (right), we set p(Light off)=0.5 during self-play\footnote{Changing
  p(Light off) adjusts the seperation between the self-play and target
tasks. For a systematic evaluation of this, please see \append{bias}
.} and evaluate the repeat form of self-play, alongside two baselines: (i)
target task only training (i.e. no self-play) and (ii) self-play with
a random policy for Alice.  With self-play, the agent 
succeeds quickly while target task-only training takes much
longer\footnote{Training was stopped for all methods except
  target-only at $5\times 10^6$ episodes.}.
\fig{4} shows details of a single training run, demonstrating how Alice and Bob
automatically build a curriculum between themselves though self-play.

\subsection{RLLab: Mountain Car}
We applied our approach to the Mountain Car task in RLLab. Here the
agent controls a car trapped in a 1-D valley. It must learn to build
momentum by alternately moving to the left and right, climbing higher
up the valley walls until it is able to
escape. Although the problem is presented as continuous, we discretize the 1-D action
space into 5 bins (uniformly sized) enabling us to use discrete policy gradient, as above.
We also added a secondary action head with binary actions to be used as STOP action.
An observation of state $s_t$ consists of the location and speed of the car.

As in \cite{Houthooft16,Tang16}, a reward of +1 is given only when the 
car succeeds in climbing the hill. In self-play, 
Bob succeeds if $\|s_b - s_a\| < 0.2$, where $s_a$ and $s_b$ are the final states (location and velocity of the car)
of Alice and Bob respectively.

The nature of the environment makes it highly asymmetric from Alice
and Bob's point of view, since it is far easier to coast down the hill
to the starting point that it is to climb up it. Hence
we exclusively use the reset form of self-play. In \fig{swimmer}~(left), we
compare this to current state-of-the-art methods, namely VIME \citep{Houthooft16}
and SimHash \citep{Tang16}. Our approach (blue) performs
comparably to both of these. We also tried using policy gradient
directly on the target task samples, but it was unable to solve the
problem. 

\subsection{RLLab: SwimmerGather}
We also applied our approach to the SwimmerGather
task in RLLab (which uses the Mujoco~\citep{mujoco} simulator), 
where the agent controls a worm with two
flexible joints, swimming in a 2D viscous fluid. 
In the target task, the agent gets reward +1 for
eating green apples and -1 for touching red bombs, which are not present during self-play.
Thus the self-play task and target tasks are different:
in the former, the worm just swims around but in the latter it must
learn to swim towards green apples and away from the red bombs. 

The observation state consists of a 13-dimensional vector describing
location and joint angles of the worm, and a 20 dimensional vector for
sensing nearby objects.  The worm takes two real values as an action,
each controlling one joint.  We add a secondary action head to our
models to handle the 2nd joint, and a third binary action head for STOP action.  As in the mountain car, we discretize the
output space (each joint is given 9 uniformly sized bins) to allow the
use of discrete policy gradients.

\begin{figure}[h!]
\begin{center}
\mbox{
\includegraphics[width=0.45\columnwidth]{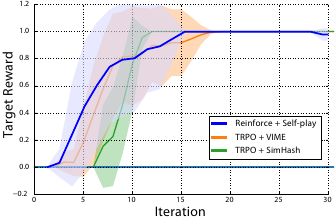}
\hspace{6mm}
\includegraphics[width=0.45\columnwidth]{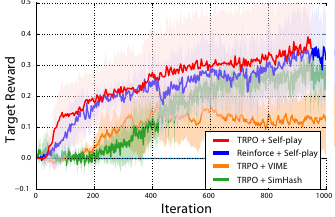}
}
\vspace{-3mm}
\caption{Evaluation on MountainCar (left) and SwimmerGather (right)
  target tasks, comparing to VIME \cite{Houthooft16} and SimHash
  \cite{Tang16} (figures adapted from \cite{Tang16}). With reversible
  self-play we are able to learn faster than the other approaches,
  although it converges to a comparable reward. Training directly on
  the target task using Reinforce without self-play resulted in total
  failure. Here 1 iteration = 5k (50k) target task steps in Mountain
  car (SwimmerGather), excluding self-play steps.}
\label{fig:swimmer}
\vspace{-8mm}
\end{center}
\end{figure}

Bob succeeds in a self-play episode when $\|l_b - l_a\| < 0.3$ where $l_a$
and $l_b$ are the final locations of Alice and Bob
respectively. \fig{swimmer}~(right) shows the target task reward as a function of
training iteration for our approach alongside state-of-the-art exploration
methods VIME \citep{Houthooft16} and
SimHash \citep{Tang16}. We demonstrate the generality of the self-play
approach by applying it to Reinforce
and also TRPO \citep{Schulman15} (see \append{swim} for details). In both cases, it enables them to
gain reward significantly earlier than 
other methods, although both converge to a similar final value to
SimHash. A video of our
worm performing the test task can be found at \url{https://goo.gl/Vsd8Js}.

\subsection{StarCraft: Training marines}
\begin{figure}[h!]
\begin{center}
\mbox{
\includegraphics[width=0.45\columnwidth]{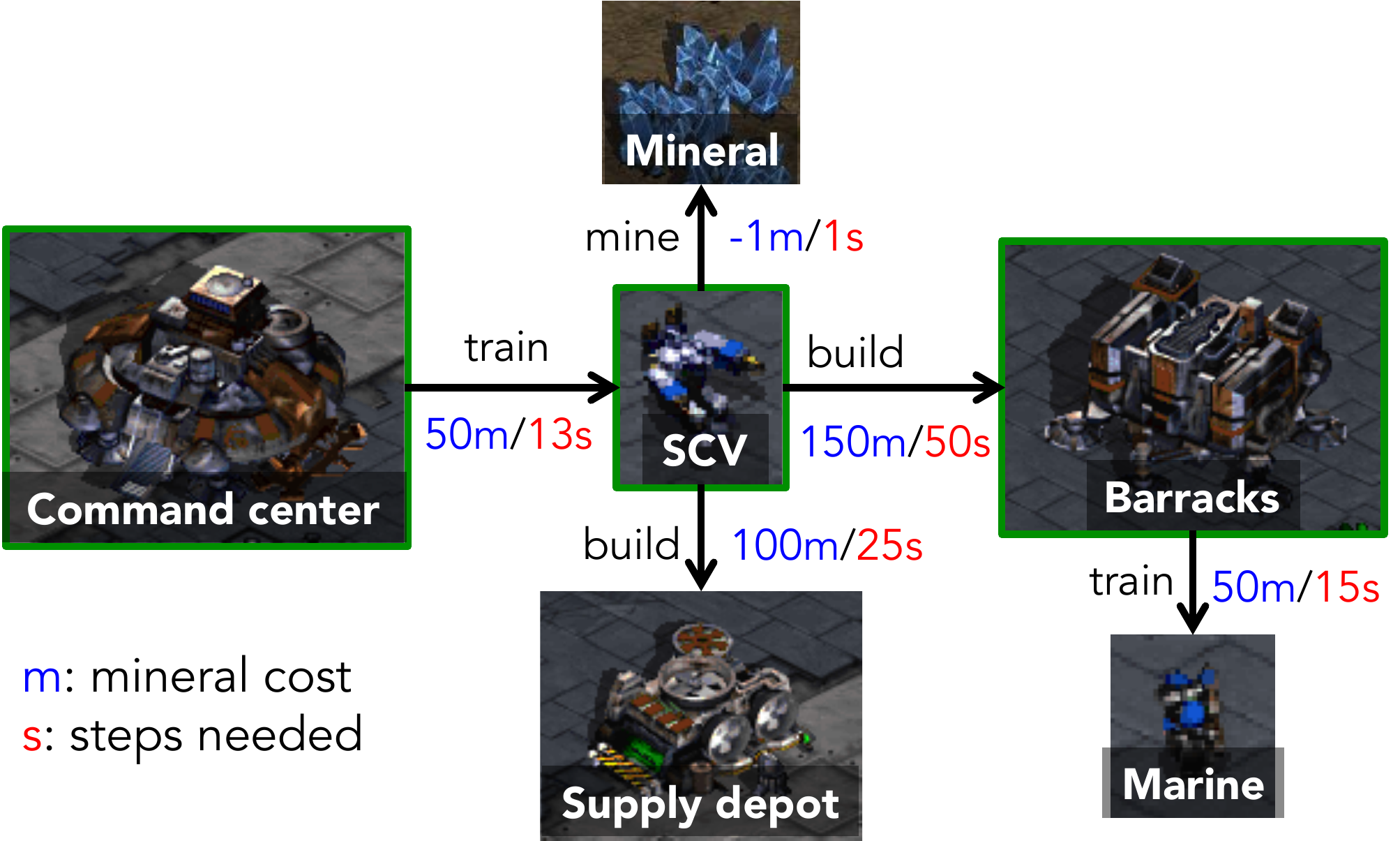}
\hspace{10mm}
\includegraphics[width=0.35\columnwidth]{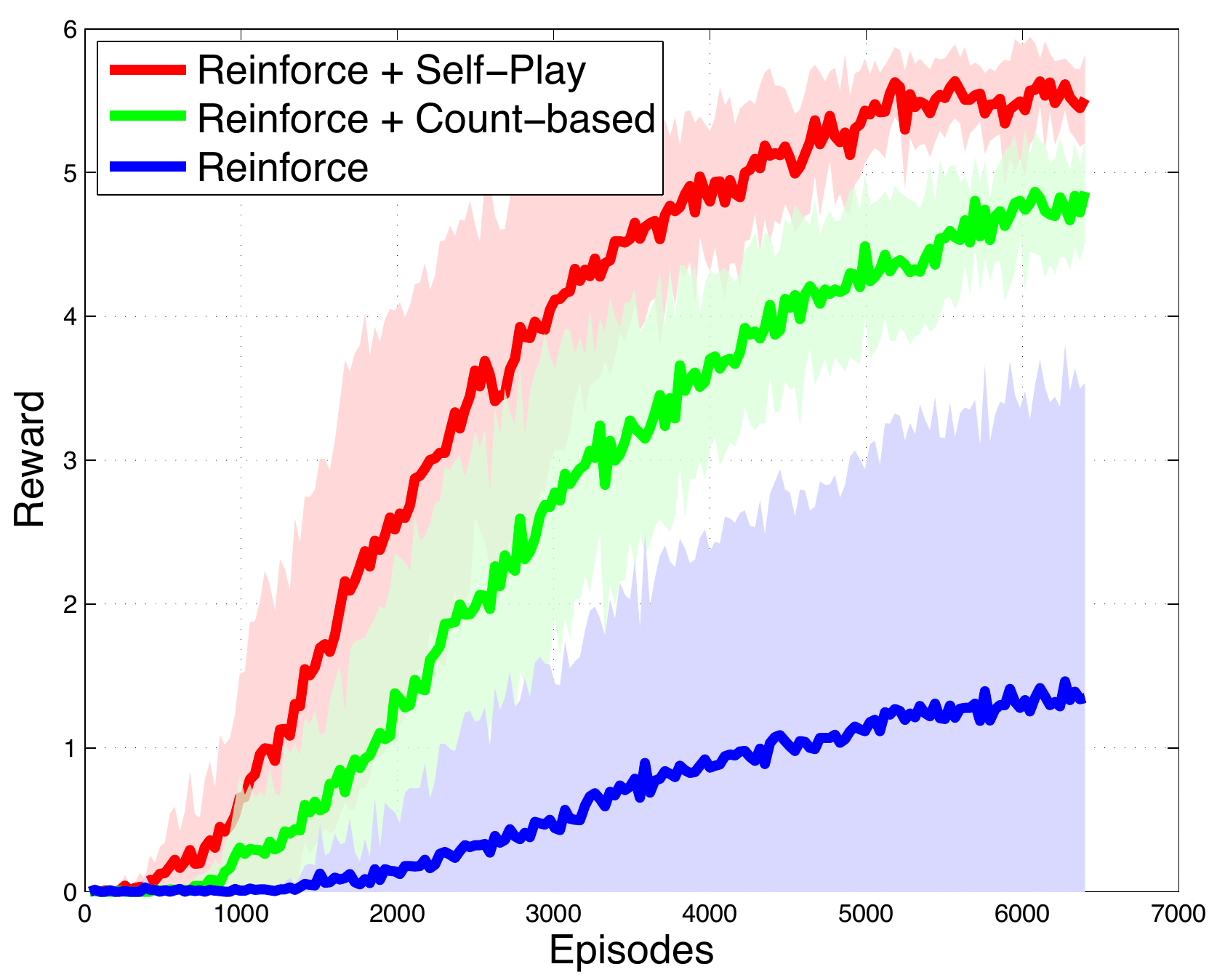}
}
\caption{
  \textbf{Left:} Different types of unit in the StarCraft environment. The arrows represent possible actions (excluding movement actions) by the unit, and corresponding numbers shows (blue) amount of minerals and (red) time steps needed to complete. The units under agent's control are outlined by a green border.
  \textbf{Right:} Plot of reward on the StarCraft sub-task of training 
  marine units vs \#target-task episodes (self-play episodes are not included),
  with and without self-play. A count-based baseline
  is also shown. Self-play greatly speeds up
  learning, and also surpasses the count-based approach at convergence.} 
\label{fig:starcraft}
\end{center}
\end{figure}

Finally, we applied our self-play approach to the same setup as
the beginning of a standard StarCraft: Brood War game~\cite{synnaeve2016torchcraft}, where an agent controls multiple units to mine, 
construct buildings, and train new units, but without enemies to fight. The environment starts
with 4 workers units (Terran SCVs), who can move around, mine nearby minerals and construct new buildings.
In addition, the agent controls the command center, which can train new workers. 
See \fig{starcraft}~(left) for relations between different units and their actions.

The target task is to build
Marine units. To do this, an agent 
must follow a specific sequence of operations: (i) mine minerals with workers; (ii)
having accumulated sufficient mineral supply, build a barracks and
(iii) once the barracks are complete, train Marine units out of it. 
Optionally, an agent can train a new worker for faster mining, 
or build a supply depot to accommodate more units.
When the episode ends after 200 steps (little over 3 minutes), 
the agent gets rewarded +1 for each Marine it has built. 
Optimizing this task is highly complex due to several factors. First, 
the agent has to find an optimal mining pattern (concentrating on a single mineral or mining a far away mineral is inefficient). Then, it has to produce
the optimal number of workers and barrack at the right timing. In addition, 
a supply depot needs to be built when the number of units is close to the limit.

During self-play (repeat variant), Alice and Bob control the workers
and can try any combination of actions during the
episode. Since exactly matching the game state is almost impossible,
Bob's success is only based on the global state of the game,
which includes the number of units of each type (including buildings),
and accumulated mineral resource. So Bob's objective in self-play
is to make as many units and mineral as Alice in shortest possible time.
Further details are given in \append{sc}. 
\fig{starcraft}~(right) compares the Reinforce algorithm on the target task,
with and without self-play. An additional count-based exploration
baseline similar to the hallway experiment is also shown. 
It utilizes the same global game state as self-play.

\section{Discussion}
\label{sec:discuss}
In this work we described a novel method for intrinsically motivated
learning which we call asymmetric self-play. Despite the method's
conceptual simplicity, we have seen that it can be effective in both
discrete and continuous input settings with function approximation,
for encouraging exploration and automatically generating
curriculums. On the challenging benchmarks we consider, our approach is
at least as good as state-of-the-art
RL methods that incorporate an incentive for exploration, despite
being based on very different principles. Furthermore,
it is possible show theoretically that in simple environments, using
asymmetric self-play with reward functions from \eqref{eq:rb} and
\eqref{eq:ra}, optimal agents can transit between any pair of
reachable states as efficiently as possible. 
The code for our approach can be found at \url{http://cims.nyu.edu/~sainbar/selfplay}.


\bibliography{paper}

\begin{thebibliography}{33}
\providecommand{\natexlab}[1]{#1}
\providecommand{\url}[1]{\texttt{#1}}
\expandafter\ifx\csname urlstyle\endcsname\relax
  \providecommand{\doi}[1]{doi: #1}\else
  \providecommand{\doi}{doi: \begingroup \urlstyle{rm}\Url}\fi

\bibitem[Andrychowicz et~al.(2017)Andrychowicz, Crow, Ray, Schneider, Fong,
  Welinder, McGrew, Tobin, Abbeel, and Zaremba]{Andrychowicz17}
Marcin Andrychowicz, Dwight Crow, Alex~K Ray, Jonas Schneider, Rachel Fong,
  Peter Welinder, Bob McGrew, Josh Tobin, Pieter Abbeel, and Wojciech Zaremba.
\newblock Hindsight experience replay.
\newblock In \emph{NIPS}, 2017.

\bibitem[Baranes \& Oudeyer(2013)Baranes and Oudeyer]{Baranes13}
A.~Baranes and P-Y. Oudeyer.
\newblock Active learning of inverse models with intrinsically motivated goal
  exploration in robots.
\newblock \emph{Robotics and Autonomous Systems}, 61\penalty0 (1):\penalty0
  49--73, 2013.

\bibitem[Barto(2013)]{Barto2013}
Andrew~G. Barto.
\newblock \emph{Intrinsic Motivation and Reinforcement Learning}, pp.\  17--47.
\newblock Springer Berlin Heidelberg, 2013.

\bibitem[Bellemare et~al.(2016)Bellemare, Srinivasan, Ostrovski, Schaul,
  Saxton, and Munos]{unifying_exploration}
Marc~G. Bellemare, Sriram Srinivasan, Georg Ostrovski, Tom Schaul, David
  Saxton, and R{\'{e}}mi Munos.
\newblock Unifying count-based exploration and intrinsic motivation.
\newblock In \emph{NIPS}, pp.\  1471--1479, 2016.

\bibitem[Bengio et~al.(2009)Bengio, Louradour, Collobert, and
  Weston]{bengio2009curriculum}
Yoshua Bengio, J{\'e}r{\^o}me Louradour, Ronan Collobert, and Jason Weston.
\newblock Curriculum learning.
\newblock In \emph{ICML}, pp.\  41--48, 2009.

\bibitem[Duan et~al.(2016)Duan, Chen, Houthooft, Schulman, and
  Abbeel]{duan2016benchmarking}
Yan Duan, Xi~Chen, Rein Houthooft, John Schulman, and Pieter Abbeel.
\newblock Benchmarking deep reinforcement learning for continuous control.
\newblock In \emph{ICML}, 2016.

\bibitem[Florensa et~al.(2017)Florensa, Held, Wulfmeier, Zhang, and
  Abbeel]{Florensa17a}
Carlos Florensa, David Held, Markus Wulfmeier, Michael Zhang, and Pieter
  Abbeel.
\newblock Reverse curriculum generation for reinforcement learning.
\newblock In \emph{CoRL}, 2017.

\bibitem[Goodfellow et~al.(2014)Goodfellow, Pouget-Abadie, Mirza, Xu,
  Warde-Farley, Ozair, Courville, and Bengio]{goodfellow2014generative}
Ian Goodfellow, Jean Pouget-Abadie, Mehdi Mirza, Bing Xu, David Warde-Farley,
  Sherjil Ozair, Aaron Courville, and Yoshua Bengio.
\newblock Generative adversarial nets.
\newblock In \emph{NIPS}, pp.\  2672--2680, 2014.

\bibitem[Held et~al.(2017)Held, Geng, Florensa, and Abbeel]{Held17}
David Held, Xinyang Geng, Carlos Florensa, and Pieter Abbeel.
\newblock Automatic goal generation for reinforcement learning agents.
\newblock \emph{CoRR}, abs/1705.06366, 2017.

\bibitem[Houthooft et~al.(2016)Houthooft, Chen, Duan, Schulman, Turck, and
  Abbeel]{Houthooft16}
Rein Houthooft, Xi~Chen, Yan Duan, John Schulman, Filip~De Turck, and Pieter
  Abbeel.
\newblock Curiosity-driven exploration in deep reinforcement learning via
  bayesian neural networks.
\newblock \emph{arXiv 1605.09674}, 2016.

\bibitem[Klyubin et~al.(2005)Klyubin, Polani, and Nehaniv]{empowerment}
Alexander~S. Klyubin, Daniel Polani, and Chrystopher~L. Nehaniv.
\newblock Empowerment: a universal agent-centric measure of control.
\newblock In \emph{Proceedings of the {IEEE} Congress on Evolutionary
  Computation, {CEC}}, pp.\  128--135, 2005.

\bibitem[Kumar et~al.(2010)Kumar, Packer, and Koller]{Kumar10}
M.~P. Kumar, Benjamin Packer, and Daphne Koller.
\newblock Self-paced learning for latent variable models.
\newblock In \emph{NIPS}. 2010.

\bibitem[Li et~al.(2017)Li, Monroe, Shi, Jean, Ritter, and Jurafsky]{Li17}
Jiwei Li, Will Monroe, Tianlin Shi, S{\'e}bastien Jean, Alan Ritter, and Daniel
  Jurafsky.
\newblock Adversarial learning for neural dialogue generation.
\newblock In \emph{EMNLP}, 2017.

\bibitem[Lopes et~al.(2012)Lopes, Lang, Toussaint, and
  Oudeyer]{model_based_exploration}
Manuel Lopes, Tobias Lang, Marc Toussaint, and Pierre{-}Yves Oudeyer.
\newblock Exploration in model-based reinforcement learning by empirically
  estimating learning progress.
\newblock In \emph{NIPS}, pp.\  206--214, 2012.

\bibitem[Mescheder et~al.(2017)Mescheder, Nowozin, and Geiger]{Mescheder17}
Lars~M. Mescheder, Sebastian Nowozin, and Andreas Geiger.
\newblock Adversarial variational bayes: Unifying variational autoencoders and
  generative adversarial networks.
\newblock In \emph{ICML}, 2017.

\bibitem[Pathak et~al.(2017)Pathak, Agrawal, Efros, and
  Darrell]{pathakICMl17curiosity}
Deepak Pathak, Pulkit Agrawal, Alexei~A. Efros, and Trevor Darrell.
\newblock Curiosity-driven exploration by self-supervised prediction.
\newblock In \emph{ICML}, 2017.

\bibitem[Pinto et~al.(2017)Pinto, Davidson, Sukthankar, and Gupta]{Pinto17}
Lerrel Pinto, James Davidson, Rahul Sukthankar, and Abhinav Gupta.
\newblock Robust adversarial reinforcement learning.
\newblock In \emph{ICML}, 2017.

\bibitem[Riedmiller et~al.(2009)Riedmiller, Gabel, Hafner, and
  Lange]{riedmiller2009reinforcement}
Martin Riedmiller, Thomas Gabel, Roland Hafner, and Sascha Lange.
\newblock Reinforcement learning for robot soccer.
\newblock \emph{Autonomous Robots}, 27\penalty0 (1):\penalty0 55--73, 2009.

\bibitem[Samuel(1959)]{DBLP:journals/ibmrd/Samuel59}
Arthur~L. Samuel.
\newblock Some studies in machine learning using the game of checkers.
\newblock \emph{{IBM} Journal of Research and Development}, 3\penalty0
  (3):\penalty0 210--229, 1959.

\bibitem[Schaul et~al.(2015)Schaul, Horgan, Gregor, and Silver]{uvfa}
Tom Schaul, Dan Horgan, Karol Gregor, and David Silver.
\newblock Universal value function approximators.
\newblock In \emph{ICML}, pp.\  1312--1320, 2015.

\bibitem[Schmidhuber(1991)]{Schmidhuber-ijcnn-91}
J.~Schmidhuber.
\newblock {Curious model-building control systems}.
\newblock In \emph{Proc. Int. J. Conf. Neural Networks}, pp.\  1458--1463. IEEE
  Press, 1991.

\bibitem[Schulman et~al.(2015)Schulman, Levine, Abbeel, Jordan, and
  Moritz]{Schulman15}
John Schulman, Sergey Levine, Pieter Abbeel, Michael~I. Jordan, and Philipp
  Moritz.
\newblock Trust region policy optimization.
\newblock In \emph{ICML}, 2015.

\bibitem[Silver et~al.(2016)Silver, Huang, Maddison, Guez, Sifre, van~den
  Driessche, Schrittwieser, Antonoglou, Panneershelvam, Lanctot, Dieleman,
  Grewe, Nham, Kalchbrenner, Sutskever, Lillicrap, Leach, Kavukcuoglu, Graepel,
  and Hassabis]{Silver_2016_go}
David Silver, Aja Huang, Chris~J. Maddison, Arthur Guez, Laurent Sifre, George
  van~den Driessche, Julian Schrittwieser, Ioannis Antonoglou, Veda
  Panneershelvam, Marc Lanctot, Sander Dieleman, Dominik Grewe, John Nham, Nal
  Kalchbrenner, Ilya Sutskever, Timothy Lillicrap, Madeleine Leach, Koray
  Kavukcuoglu, Thore Graepel, and Demis Hassabis.
\newblock Mastering the game of {Go} with deep neural networks and tree search.
\newblock \emph{Nature}, 529\penalty0 (7587):\penalty0 484--489, January 2016.

\bibitem[Singh et~al.(2004)Singh, Barto, and
  Chentanez]{intrinsically_motivated_RL}
Satinder~P. Singh, Andrew~G. Barto, and Nuttapong Chentanez.
\newblock Intrinsically motivated reinforcement learning.
\newblock In \emph{NIPS}, pp.\  1281--1288, 2004.

\bibitem[Strehl \& Littman(2008)Strehl and Littman]{mbie}
Alexander~L. Strehl and Michael~L. Littman.
\newblock An analysis of model-based interval estimation for markov decision
  processes.
\newblock \emph{J. Comput. Syst. Sci.}, 74\penalty0 (8):\penalty0 1309--1331,
  2008.

\bibitem[Sukhbaatar et~al.(2015)Sukhbaatar, Szlam, Synnaeve, Chintala, and
  Fergus]{mazebase}
Sainbayar Sukhbaatar, Arthur Szlam, Gabriel Synnaeve, Soumith Chintala, and Rob
  Fergus.
\newblock Mazebase: {A} sandbox for learning from games.
\newblock \emph{arXiv 1511.07401}, 2015.

\bibitem[Sutton et~al.(2011)Sutton, Modayil, Delp, Degris, Pilarski, White, and
  Precup]{horde}
Richard~S. Sutton, Joseph Modayil, Michael Delp, Thomas Degris, Patrick~M.
  Pilarski, Adam White, and Doina Precup.
\newblock Horde: A scalable real-time architecture for learning knowledge from
  unsupervised sensorimotor interaction.
\newblock In \emph{AAMAS '11}, pp.\  761--768, 2011.

\bibitem[Synnaeve et~al.(2016)Synnaeve, Nardelli, Auvolat, Chintala, Lacroix,
  Lin, Richoux, and Usunier]{synnaeve2016torchcraft}
Gabriel Synnaeve, Nantas Nardelli, Alex Auvolat, Soumith Chintala, Timoth{\'e}e
  Lacroix, Zeming Lin, Florian Richoux, and Nicolas Usunier.
\newblock Torchcraft: a library for machine learning research on real-time
  strategy games.
\newblock \emph{arXiv preprint arXiv:1611.00625}, 2016.

\bibitem[Tang et~al.(2017)Tang, Houthooft, Foote, Stooke, Chen, Duan, Schulman,
  Turck, and Abbeel]{Tang16}
Haoran Tang, Rein Houthooft, Davis Foote, Adam Stooke, Xi~Chen, Yan Duan, John
  Schulman, Filip~De Turck, and Pieter Abbeel.
\newblock \#exploration: A study of count-based exploration for deep
  reinforcement learning.
\newblock In \emph{NIPS}, 2017.

\bibitem[Tesauro(1995)]{DBLP:journals/cacm/Tesauro95}
Gerald Tesauro.
\newblock Temporal difference learning and td-gammon.
\newblock \emph{Commun. {ACM}}, 38\penalty0 (3):\penalty0 58--68, 1995.

\bibitem[Tieleman \& Hinton(2012)Tieleman and Hinton]{rmsprop}
T.~Tieleman and G.~Hinton.
\newblock Lecture 6.5 - rmsprop, coursera: Neural networks for machine
  learning, 2012.

\bibitem[Todorov et~al.(2012)Todorov, Erez, and Tassa]{mujoco}
Emanuel Todorov, Tom Erez, and Yuval Tassa.
\newblock Mujoco: A physics engine for model-based control.
\newblock In \emph{IROS}, pp.\  5026--5033. IEEE, 2012.

\bibitem[Williams(1992)]{Williams92simplestatistical}
Ronald~J. Williams.
\newblock Simple statistical gradient-following algorithms for connectionist
  reinforcement learning.
\newblock In \emph{Machine Learning}, pp.\  229--256, 1992.

\end{thebibliography}
\bibliographystyle{iclr2018_conference}

\appendix

\section{Pseudo Code}
\label{app:code}

Algorithm~\ref{alg:selfplay} and \ref{alg:target} are the pseudo codes for training an agent on self-play and target task episodes.

\begin{algorithm}[!h]
\begin{algorithmic}
\small
\Function{SelfPlayEpisode(\textcolor{red}{Reverse}/\textcolor{blue}{Repeat},$t_\text{Max},\theta_A,\theta_B$)}{}
\State $t_A \leftarrow 0$
\State $s_0 \leftarrow$ env.observe()
\State \textcolor{red}{$s^* \leftarrow s_0$} 
\While{True}
       \State \# Alice's turn
       \State $t_A \leftarrow t_A + 1 $
       \State $s \leftarrow$ env.observe()
       \State $a \leftarrow$ $\pi_A(s, s_0) = f(s, s_0, \theta_A)$
    \If{$a =$ {\sc stop} {\bf or} $t_A \ge t_\text{Max}$}
        \State \textcolor{blue}{$s^* \leftarrow s$}
        \State \textcolor{blue}{env.reset()}
    	\State \textbf{break}
    \EndIf
    \State env.act($a$) 
\EndWhile

\State $t_B \leftarrow 0$
\While{True}
       \State \# Bob's turn
  \State $s \leftarrow$ env.observe()
    \If{$s = s^*$ {\bf or} $t_A+t_B \ge t_\text{Max}$} 
      \State \textbf{break}
    \EndIf
	\State $t_B \leftarrow t_B + 1 $
    \State $a \leftarrow$ $\pi_B(s, s^*) = f(s, s^*,\theta_B)$
    \State env.act($a$)
\EndWhile
\State $R_A \leftarrow \gamma \max(0,t_B -  t_A)$
\State $R_B \leftarrow -\gamma t_B$
\State policy.update($R_A, \theta_A$)
\State policy.update($R_B, \theta_B$)
\State {\bf return} 
\EndFunction
\end{algorithmic}
\caption{Pseudo code for training an agent on a self-play episode}
\label{alg:selfplay}
\end{algorithm}

\begin{algorithm}[!h]
\begin{algorithmic}
\small
\Function{TargetTaskEpisode($t_\text{Max},\theta_B$)}{}
\State $t \leftarrow 0$
\State $R \leftarrow 0$
\While{True}
    \State $t \leftarrow t + 1 $
       \State $s \leftarrow$ env.observe()
       \State $a \leftarrow$ $\pi_B(s, \emptyset) = f(s, \emptyset, \theta_B)$
    \If{env.done() {\bf or} $t \ge t_\text{Max}$}
      \State \textbf{break}
    \EndIf
    \State env.act($a$)
    \State $R = R + $ env.reward() 
\EndWhile

\State policy.update($R, \theta_B$)
\State {\bf return} 
\EndFunction
\end{algorithmic}
\caption{Pseudo code for training an agent on a target task episode}
\label{alg:target}
\end{algorithm}

\section{Hyperparameters used in the experiments}
\label{app:hyper}
For the experiments with neural networks, all parameters are randomly initialized from $\mathcal{N}(0, 0.2)$. The Hyperparameters of RMSProp are set to 0.97 and $1e-6$.
The other hyperparameter values used in the experiments are shown in \tab{hyper}. In some cases, we used different parameters for self-play and target task episodes. 
Entropy regularization is implemented as an additional cost maximizing the entropy of the softmax layer.
In the StarCraft, skipping 23 frames roughly matches to one action per second.
\begin{table}[h]
\centering
\begin{tabular}{|l||c|c|c|c|c|} \hline
Hyperparameter                & Long          & Mazebase  & Mountain      & Swimmer         & StarCraft   \\
name                          & Hallway       &           &          Car  &        Gather   &    \\ \hline \hline
Learning rate                 & 0.1           & 0.003     & 0.003         & 0.003           & 0.003       \\ \hline
Batch size                    & 16            & 256       & 128           & 256             & 32          \\ \hline
Max steps of                  & 30            & 80        & 500           & TT: 166  & 200          \\
episode ($t_\text{max}$)      &               &           &               & SP: 200    &           \\ \hline
Entropy                       & 0             & 0.003     & 0.003         & TT: 0  & TT: 0 \\
regularization                &               &           &               & SP:  0.003 & SP:  0.003   \\ \hline
Self-play reward scale ($\gamma$)       & 0.033             & 0.1       & 0.01          & 0.01            & 0.01       \\ \hline
Self-play percentage          & -             & 20\%      & 1\%           & 10\%            & 10\%       \\ \hline
Self-play mode                & Reverse       & Both      & Repeat        & Reverse          & Repeat       \\ \hline
Frame skip                    & 0             & 0         & 0             & 150 \footnotemark  & 23         \\ \hline
\end{tabular}
\caption{Hyperparameter values used in experiments. TT=target task, SP=self-play}
\label{tab:hyper}
\end{table}
\footnotetext{Experiments in VIME and SimHash papers skip 50 frames, but we matched the total number of frames in an episode by reducing the number of steps.}

\section{Mazebase}
The agent has full visibility of the maze when the light is on.
If light is off, the agent can only see the light switch.
In self-play, Bob does not need to worry about things that are invisible to him.
For example, if Alice started with light ``off'' in reverse self-play, 
Bob does not need to match the state of the door, because it would be invisible to him
when the light is off.

In the target task, the agent and the goal are always placed on opposite
sides of the wall. Also, the light and key switches are placed on the
same side as the agent, but the light is always off and the door is closed
initially. Therefore, in order to succeed, the agent has to turn on
the light, toggle the key switch to open the door, pass through it, and reach the goal flag.

Both Alice and Bob's policies are modeled by a fully-connected neural network
with two hidden layers each with 100 and 50 units (with tanh non-linearities)
respectively. The encoder into each of the networks takes a bag of words over
(objects, locations); that is, there is a separate word in the lookup
table for each (object, location) pair. Action probabilities are output by a linear layer
followed by a softmax. 

\subsection{Biasing for or against self-play}
\label{app:bias}
The effectiveness of our approach depends in part on the similarity between the
self-play and target tasks. One way to explore this in our environment
is to vary the probability of the
light being off initially during self-play episodes\footnote{The
  initial state of the light should dramatically change the behavior
  of the agent: if it is on then agent can directly proceed to the
  key.}. Note that the light
is always off in the target task; if the light is usually on at the 
start of Alice's turn in reverse, for example, she will learn to turn it off, and then Bob will
be biased to turn it back on.  On the other hand, if the light is usually off
at the start of Alice's turn in reverse, Bob is strongly biased against turning the
light on, and so the test task becomes especially hard. Thus changing this probability gives
us some way to adjust the similarity between the two tasks.   

\begin{figure}[h!]
\begin{center}
\mbox{
\includegraphics[width=0.4\columnwidth]{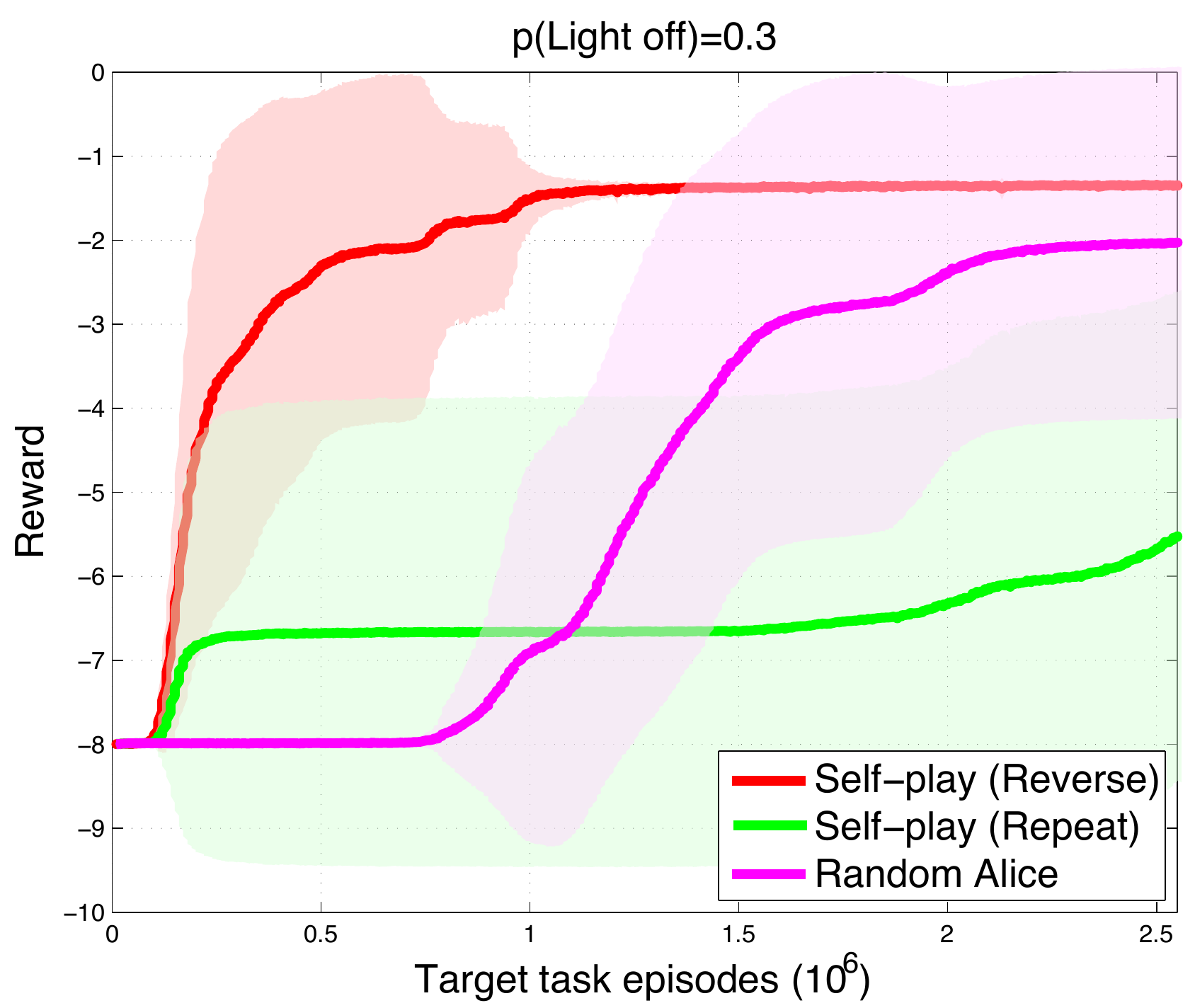}
\hspace{6mm}
\includegraphics[width=0.4\columnwidth]{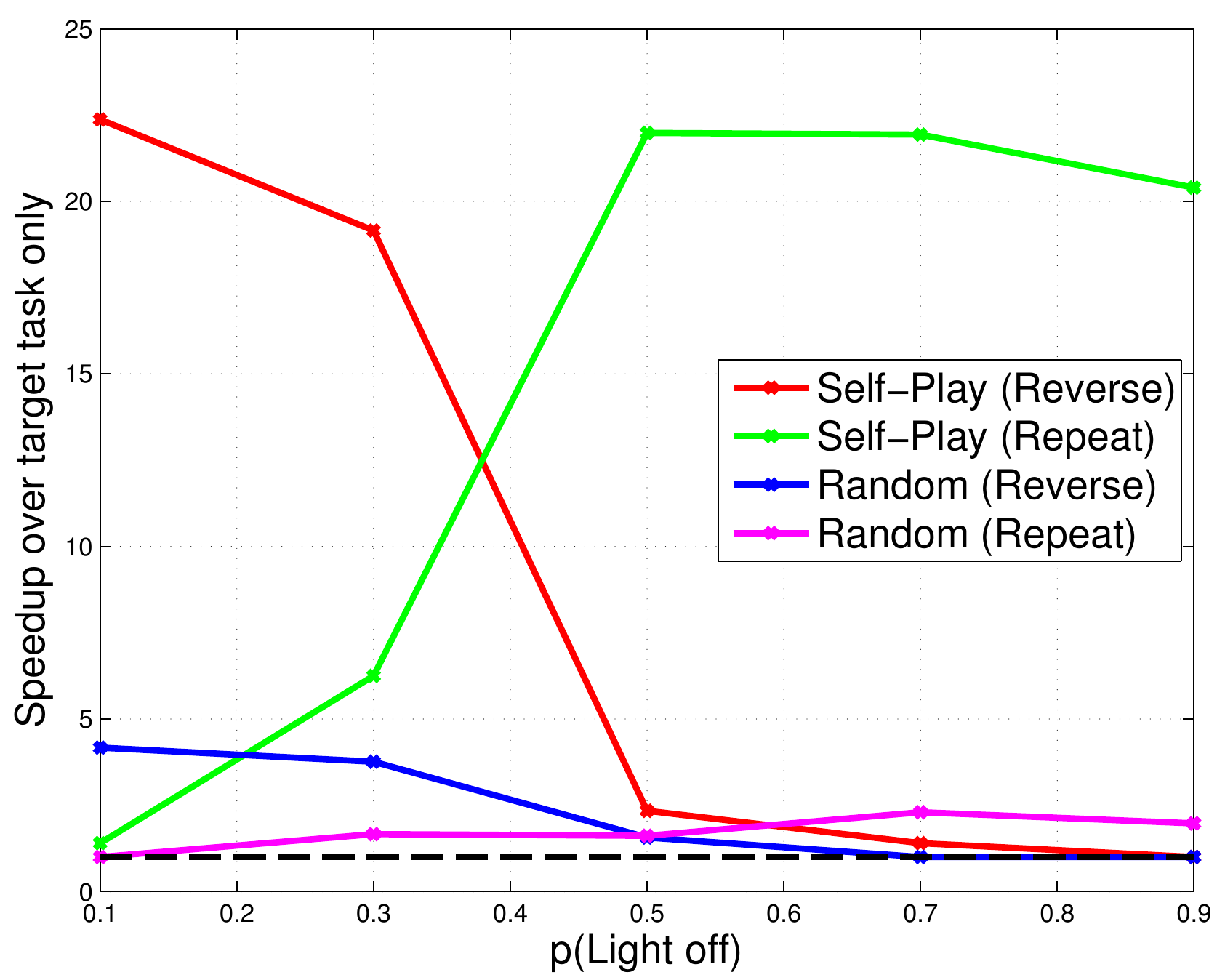}
}
\caption{Left: The performance of self-play when p(Light off) set to 0.3. Here the
reverse form of self-play works well (more details in the text).
  Right: Reduction in target task episodes relative to training purely
  on the target-task as the distance between self-play and the target
  task varies (for runs where the reward goes above -2 on the
  Mazebase task -- unsuccessful runs are given a unity speed-up factor).
  The $y$ axis is the speedup, and $x$ axis is p(Light off). For
  reverse self-play, the low p(Light off) corresponds to having
  self-play and target tasks be similar to one another, while the
  opposite applies to repeat self-play. For both forms, significant
  speedups are achieved when self-play is similar to the target tasks,
but the effect diminishes when self-play is biased against the target
task.}
\label{fig:mazebase_speedup}
\end{center}
\end{figure}

\fig{mazebase_speedup} (left) shows what happens when p(Light off)=0.3. Here reverse
self-play works well, but repeat self-play does poorly.  As discussed above, this
flipping, relative to the previous experiment, can be explained as follows:
low p(Light off) means that Bob's task in reverse self-play will typically involve
returning the light to the on position (irrespective of how Alice left
it), the same function that must be performed in the target task. The
opposite situation applies for repeat self-play, where Bob needs to 
encounter the light typically in the off position to help him with the
test task.

In \fig{mazebase_speedup} (right) we systematically vary p(Light off) between
0.1 and 0.9. The y-axis shows the speed-up (reduction in target task
episodes) relative to training purely on the target-task for runs
where the reward goes above -2. Unsuccessful runs are given a unity
speed-up factor. The curves show that when the self-play task is not
biased against the target task it can help significantly. 

\section{SwimmerGather Experiment}
\label{app:swim}

\begin{figure}[h]
\begin{center}
\includegraphics[width=\columnwidth]{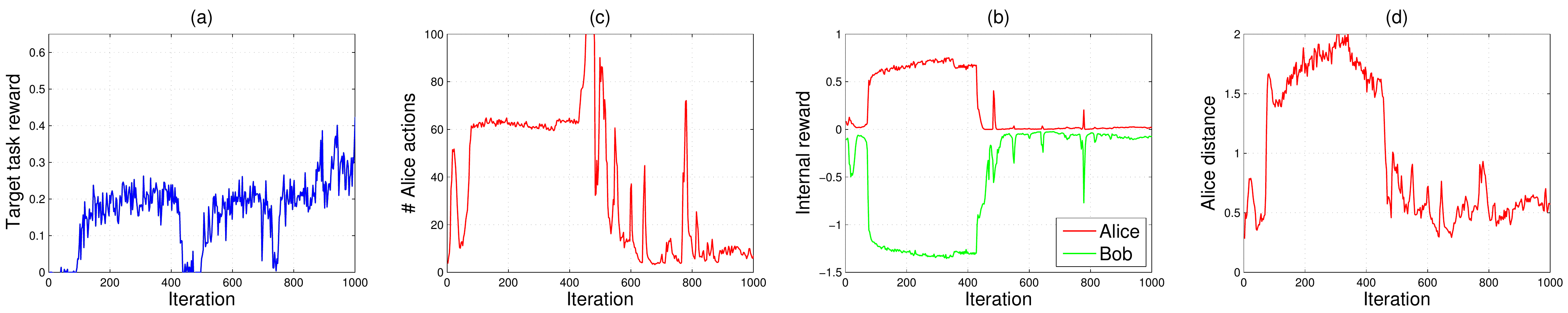}
\vspace{-8mm}
\caption{A single SwimmerGather training run. (a): Rewards on target
  task. (b): Rewards from reversible self-play.  (c): 
The number of actions taken by Alice. (d): Distance that Alice travels
before switching to Bob.} 
\label{fig:swimmer2}
\end{center}
\end{figure}

\begin{figure}[h]
\begin{center}
\includegraphics[width=0.75\columnwidth]{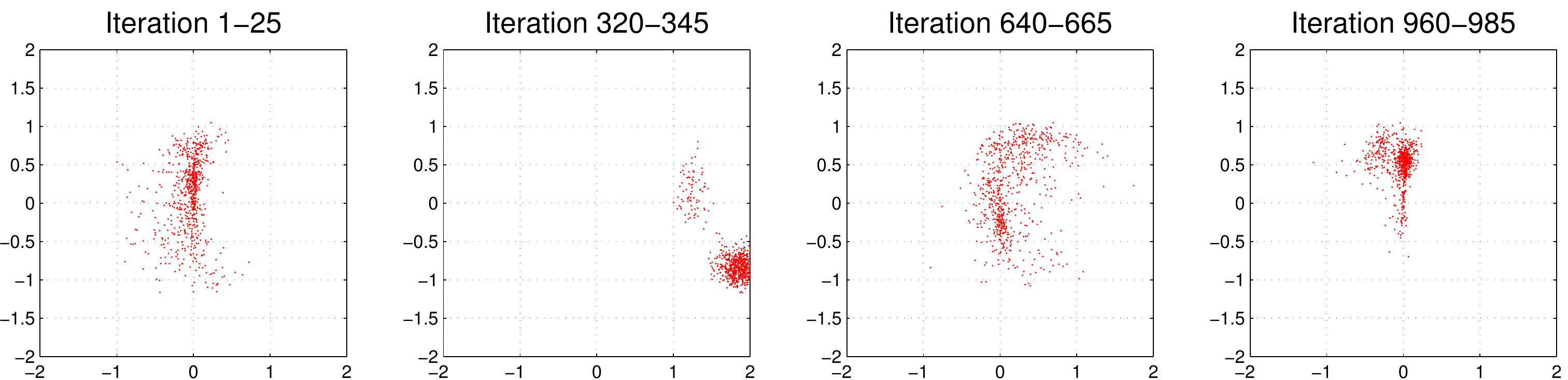}
\caption{Plot of Alice's location at time of {\sc STOP} action
  for the SwimmerGather training run shown in
  \fig{swimmer2}, for different stages of training. Note how Alice's
  distribution changes as Bob learns to solve her tasks.} 
\label{fig:scatter}
\end{center}
\end{figure}

In \fig{swimmer2} shows details of a single training run. The changes
in Alice's behavior, observed in \fig{swimmer2}(c) and (d), correlate
with Alice and Bob's reward (\fig{swimmer2}(b)) and, initially at
least, to the reward on the test target (\fig{swimmer2}(a)).
In \fig{scatter} we visualize for a single training run the locations where Alice hands
over to Bob at different stages of training, showing how the
distribution varies. 

In the TRPO experiment, we used step size 0.01 and damping coefficient 0.1.
The batch consists of 50,000 steps, of which 25\% 
comes from target task episodes, while the remaining 75\% is from self-play.
The self-play reward scale $\gamma$ set to 0.005.
We used two separate network for actor and critic, and the critic network has 
L2 weight regularization with coefficient of $1e-5$.

\section{StarCraft Experiment}
\label{app:sc}
We call units that perform action as active unit. This includes
worker units (SCVs), the command center, and barrack. 
The agent controls multiple active units in parallel. 
At each time step, an action of $i$'th unit is output by
\[ a_t^i = \pi(s_t^i, \hat{s}_t), \]
where $s_t^i$ is a unit specific local observation, and $\hat{s}_t$ is an global observation.
With $s_t^i$, a unit can see the 64x64 area around it with a resolution of 4 (unit's type is also visible).
The global observation contains the number of units and accumulated minerals in the game
\[\hat{s}_t = \{ \lfloor N_\text{ore}/25 \rfloor, N_\text{SCV}, N_\text{Barrack}, N_\text{SupplyDepot}, N_\text{Marines} \}.\]

In self-play, Bob perceives only the global observation of his target state
\[\pi_B(s_t^i, \hat{s}_t, \hat{s}^*),\]
where $\hat{s}^*$ is the final global observation of Alice. Bob will succeed only if
\[\forall i \quad \hat{s}_t[i] \ge \hat{s}^*[i] .\]

\tab{sc_action} shows the action space of different unit types controlled by the agent. The number of possible action is the same for all units since they controlled by a single model (unit type is encoded in the observation), but the meaning of actions differ according to unit type. An empty cell mean that the unit does nothing (nothing is sent to StarCraft, so the previous action persists). 

The more complexes actions ``mine minerals'', ``build a barracks'', ``build a supply depot" have the following semantics, respectively: mine the mineral closest to the current unit, build a barracks at the position of the current unit, build a supply depot on the position of the current unit.

Some actions are ignored under certain conditions: ``mining'' action is ignored if the distance to the closest mineral is greater than 12; ``switch to Bob'' is ignored if Bob is already in control; ``building'' and ``training'' actions are ignored if there is not enough resources; the actions that create a new SCV or a barracks are ignored if the number of active units is reached the limit of 10. Also ``build'' actions will be ignored if there is not enough room to build at the unit's location.

For the count-based exploration, we gave an extra reward of $\alpha /\sqrt{N(\hat{s_t})}$ at every step, where $N$ is the visit count function and $\hat{s_t}$ is a global observation. We found $\alpha=0.1$ to be works the best.

\begin{figure}[h]
\begin{center}
\includegraphics[width=0.65\columnwidth]{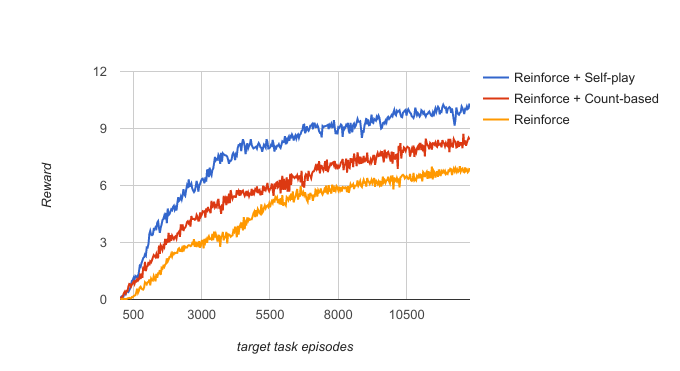}
\caption{
  Plot of reward on the StarCraft sub-task of training 
  where episode length $\tmax$ is increased to 300.} 
\label{fig:sc2}
\end{center}
\end{figure}

In \fig{sc2} we show the result of an additional experiment where we extended the length of the episode from 200 to 300, giving more time to the agent for development. The self-play still outperforms baselines methods. Note that to make more than 6 marines, an agent has to build a supply depot as well as a barracks.

\begin{table}
\centering
\begin{tabular}{|c||l|l|l|} \hline
Action ID & SCV         & Command center          & Barraks   \\ \hline \hline
 1 & move to right & train SCV & train a marine\\ \hline
 2 & move to left & switch to Bob & \\ \hline
 3 & move to top &  & \\ \hline
 4 & move to bottom & & \\ \hline
 5 & mine minerals & & \\ \hline
 6 & build a barracks & & \\ \hline
 7 & build a supply depot & & \\ \hline
\end{tabular}
\caption{Action space of different unit types in StarCraft. }
\label{tab:sc_action}

\end{table}

\section{Further Discussion}

\subsection{Meta-exploration for Alice}
\label{sec:discuss2}
 We want Alice and Bob to explore the state (or state-action) space, and we would like Bob to be exposed to many different tasks.  Because of the form of the standard reinforcement learning objective (expectation over rewards), Alice only wants to find the single hardest thing for Bob, and is not interested in the space of things that are hard for Bob.   In the fully tabular setting, with fully reversible dynamics or with resetting, and without the constraints of realistic optimization strategies, we saw in section 2.2 that this ends up forcing Bob and Alice to learn to make any state transition as efficiently as possible.  However, with more realistic optimization methods or environments, and with function approximation, Bob and Alice can get stuck in sub-optimal minima.  
 
 For example, let us follow the argument in the third paragraph of Sec.~2.2, and assume that Bob and Alice are at an equilibrium (and that we are in the tabular, finite, Markovian setting), but now we can only update Bob's and Alice's policy locally.  By this we mean that in our search for a better policy for Bob or Alice, we can only make small perturbations, as in policy gradient algorithms.   In this case, we can only guarantee that Bob runs a fast policy on challenges that Alice has non-zero probability of giving; but there is no guarantee that Alice will cover all possible challenges.    With function approximation instead of tabular policies, we cannot make any guarantees at all.  
 
 Another example with a similar outcome but different mechanism can occur using the reverse game in an environment without fully reversible dynamics.  In that case, it could be that the shortest expected number of steps to complete a challenge $(s_0,s_T)$  is longer than the reverse, and indeed, so much longer that Alice should concentrate all her energy on this challenge to maximize her rewards.  Thus there could be equilibria with Bob matching the fast policy only for a subset of challenges even if we allow non-local optimization. 
  
 The result is that Alice can end up in a policy that is not ideal for our purposes.  In figure \ref{fig:scatter}  we show the distributions of where Alice cedes control to Bob in the swimmer task.  We can see that Alice has a preferred direction.  Ideally, in this environment, Alice would be teaching Bob how to get from any state to any other efficiently; but instead, she is mostly teaching him how to move in one direction.

One possible approach to correcting this is to have multiple Alices, regularized so that they do not implement the same policy.  More generally, we can investigate objectives for Alice that encourage her to cover a wider distribution of behaviors.

\subsection{Communicating via actions}
\label{app:comm}
In this work we have limited Alice to propose tasks for Bob by doing them.  This limitation is practical and effective in restricted environments that allow resetting or are (nearly) reversible.  It allows a solution to three of the key difficulties of implementing the basic idea of ``Alice proposes tasks, Bob does them'':  parameterizing the sampling of tasks, representing and communicating the tasks, and ensuring the appropriate level of difficulty of the tasks.
Each of these is interesting in more general contexts.   In this work, the tasks have incentivized efficient transitions.   One can imagine other reward functions and task representations that incentivize discovering statistics of the states and state-transitions, for example models of their causality or temporal ordering, cluster structure.   

\end{document}